\documentclass[10pt,twocolumn,letterpaper]{article}

\usepackage{cvpr}              
\usepackage{multicol}
\usepackage{multirow}
\usepackage{bbm}
\usepackage[dvipsnames, svgnames, x11names, table]{xcolor}
\usepackage{graphicx}
\usepackage{amsmath}
\usepackage{amssymb}
\usepackage{booktabs}
\usepackage{algorithm}
\usepackage{algorithmic}
\usepackage{array}
\usepackage{url}
\usepackage{color}
\usepackage{makecell}
\usepackage[accsupp]{axessibility}

\usepackage[pagebackref,breaklinks,colorlinks]{hyperref}

\usepackage[capitalize]{cleveref}
\crefname{section}{Sec.}{Secs.}
\Crefname{section}{Section}{Sections}
\Crefname{table}{Table}{Tables}
\crefname{table}{Tab.}{Tabs.}

\def\cvprPaperID{8746} 
\def\confName{CVPR}
\def\confYear{2023}

\begin{document}

\title{Beyond Appearance: a Semantic Controllable Self-Supervised Learning Framework for Human-Centric Visual Tasks}

\author{Weihua Chen, Xianzhe Xu, Jian Jia, Hao luo, Yaohua Wang, Fan Wang, Rong Jin, Xiuyu Sun\footnotemark[2]\\
Alibaba Group\\
{\tt\small \{kugang.cwh, xianzhe.xxz, jj359864, michuan.lh, xiachen.wyh, fan.w, jinrong.jr, } \\
{\tt\small xiuyu.sxy\}@alibaba-inc.com}
}
\maketitle
\renewcommand{\thefootnote}{\fnsymbol{footnote}}
\footnotetext[2]{Corresponding Author}
\renewcommand{\thefootnote}{\arabic{footnote}}

\begin{abstract}
Human-centric visual tasks have attracted increasing research attention due to their widespread applications. 
In this paper, we aim to learn a general human representation from massive unlabeled human images which can benefit downstream human-centric tasks to the maximum extent. 
We call this method SOLIDER, a Semantic cOntrollable seLf-supervIseD lEaRning framework.  
Unlike the existing self-supervised learning methods, prior knowledge from human images is utilized in SOLIDER to build pseudo semantic labels and import more semantic information into the learned representation.
Meanwhile, we note that different downstream tasks always require different ratios of semantic information and appearance information. 
For example, human parsing requires more semantic information, while person re-identification needs more appearance information for identification purpose.
So a single learned representation cannot fit for all requirements.
To solve this problem, SOLIDER introduces a conditional network with a semantic controller. After the model is trained, users can send values to the controller to produce representations with different ratios of semantic information, which can fit different needs of downstream tasks.
Finally, SOLIDER is verified on six downstream human-centric visual tasks. It outperforms state of the arts and builds new baselines for these tasks.
The code is released in \href{https://github.com/tinyvision/SOLIDER}{https://github.com/tinyvision/SOLIDER}.
\end{abstract}
\vspace{-0.2cm}

\section{Introduction}
\label{sec:intro}
\vspace{-0.1cm}
Human-centric visual analysis plays an important role in widespread applications, such as surveillance, sports, augmented reality, and video production. Person re-identification~\cite{reidsurvey1,reidsurvey2,chen2017beyond,chen2017aaai}, attribute recognition~\cite{attrsurvey1,attrsurvey2}, person search~\cite{cuhk-sysu,prw}, pedestrian detection~\cite{detsurvey1,detsurvey2,pedestron}, human parsing~\cite{parsesurvey1,lip}, and pose estimation~\cite{posesurvey1,posesurvey2}  have achieved considerable progress in recent years. 
In another aspect, there are massive human images available in the current computer vision community. For example, even an unlabeled person re-identification dataset, LUPerson~\cite{lup,fu2022large} (\#Img$\approx$4.18M) is 4 time larger than the ImageNet dataset (\#Img$\approx$1M).  
How to use unlabeled data to build a human representation is challenging, especially when it needs to benefit various downstream tasks.

\begin{figure}[!t]
\centering
\includegraphics[width=1.0\linewidth]{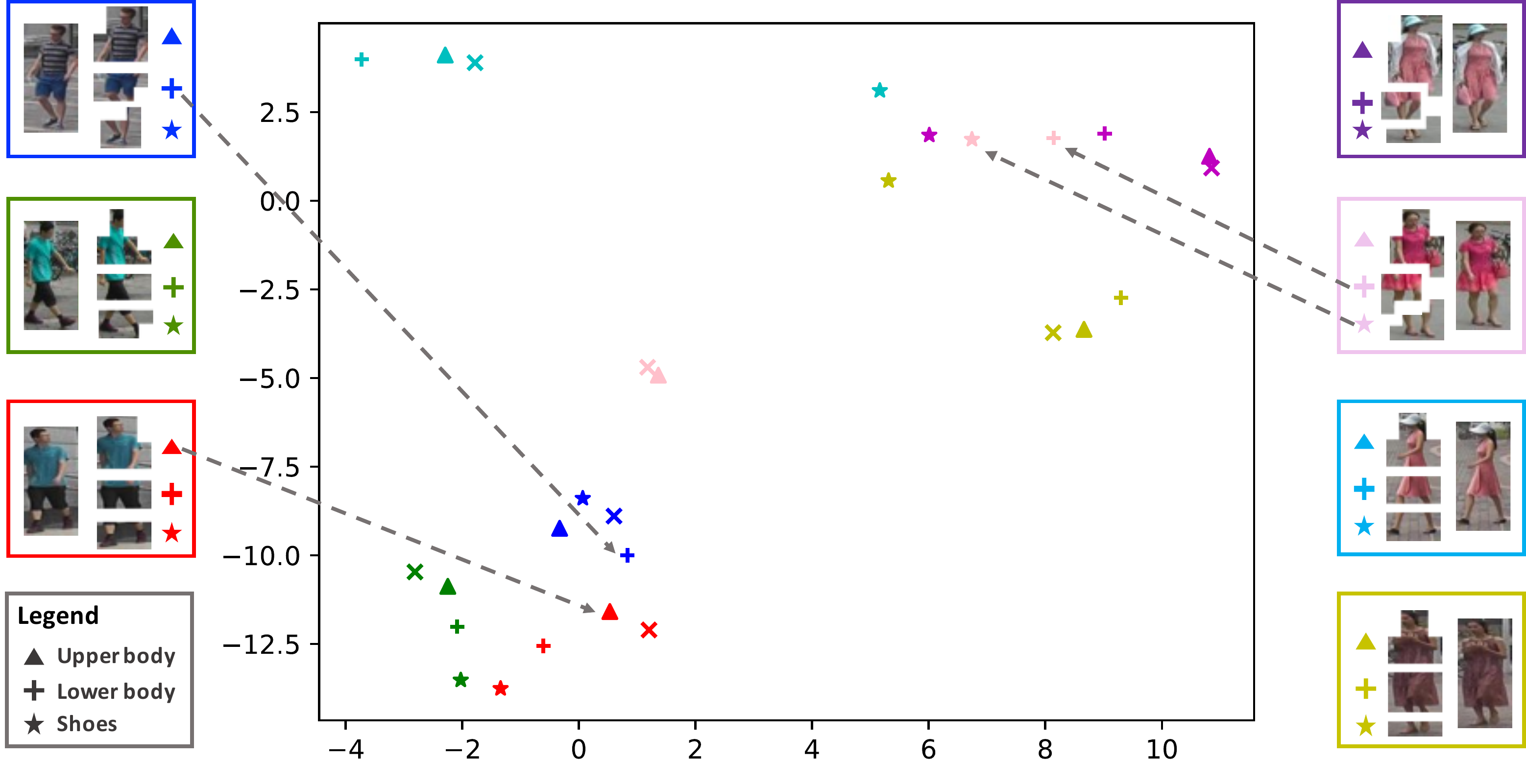}
\vspace{-0.4cm}
\caption{A representation space learned by DINO~\cite{dino}. Seven human images are represented in seven different colors. Each image is split into four parts according to their semantic regions, \ie, upper body (as $\blacktriangle$), lower body (as \textbf{+}), shoes (as $\star$) and background (as \textbf{$\times$}, not visualized to avoid distraction). It can be seen that different parts of a same person are closer to each other even they share different semantic meanings.}
\label{fig:dino_space}
\vspace{-0.1cm}
\end{figure}

Self-supervised learning has achieved great developments by using unlabeled data to learn representations. Many pretext tasks have been designed, such as contrastive learning~\cite{dino, moco, simclr} and masking image modeling~\cite{ mae, simmim, ibot, beit}. Although these methods have achieved great success in learning general image representations, there is a lack of specific design targeting human-centric tasks.

Some researchers~\cite{luo2021self,pass,yang2022unleashing} focus to extend self-supervised learning methods on human-centric visual tasks. 
They use DINO~\cite{dino} with LUPerson~\cite{lup,fu2022large} dataset to build pre-trained models for person re-identification task. 
When applying the pre-trained models to other human-centric tasks, such as human parsing and pedestrian detection, we usually get sub-optimal results. It is due to the lack of semantic information in their learned representations.  
As shown in Fig.~\ref{fig:dino_space}, in the representation space learned by DINO~\cite{dino}\footnote{MAE~\cite{mae} shares a similar phenomenon as DINO~\cite{dino}.}, 
different parts of a same person are gathered together due to their appearance continuity, no matter what semantic meanings they have.

As we've known, semantic information is as important as appearance information for human-centric visual tasks~\cite{liu2019braidnet,isp,jin2020semantics}. Therefore, we tend to train the representation with more semantic information to extend the representation to different downstream human-centric visual tasks. 
In this paper,  a Semantic cOntrollable seLf-supervIseD lEaRning framework (SOLIDER) is proposed.
In SOLIDER, we take advantage of prior knowledge from human images to discover semantic information, which can produce pseudo semantic labels for every token. And a token-level semantic classification pretext task is imported and supervised by these pseudo labels.
With the new pretext task, we can train representation with stronger semantic information. 

During the usage of our trained representation on downstream tasks, we find that even though semantic information and appearance information are both important, different downstream tasks require different ratios of them. 
Adjusting their ratio in the representation would lead to a better performance in downstream tasks. However, as long as the pretext task is trained, the representation can not be changed in current self-supervised learning methods. 
Different from previous methods, we design SOLIDER as a conditional network involving a semantic controller. 
The controller takes a value as input and produces a latent representation.
In the usage of the pre-trained model from SOLIDER, we send a value (indicting the ratio of semantic information in the representation) to the controller which can adjust the model and output a representation with the required ratio.

In summary, our paper makes four contributions:

1) A general human representation is learned in this paper, which is used as a better pre-trained model benefiting to downstream human-centric visual tasks.

2) A semantic controllable self-supervised learning framework (SOLIDER) is proposed. It takes advantages of prior knowledge in human images to produce pseudo semantic labels, and utilize it to train the human representation with more semantic information.

3) A semantic controller is designed in SOLIDER. With the controller, the pre-trained model can generate representations with various degrees of semantic information that can meet different needs of downstream tasks.

4) The effectiveness of the SOLIDER representation is verified on six downstream human-centric tasks. 
We believe this paper can promote the development of these human-centric tasks in computer vision community.

\begin{figure*}[!t]
\centering
\includegraphics[width=1.0\linewidth]{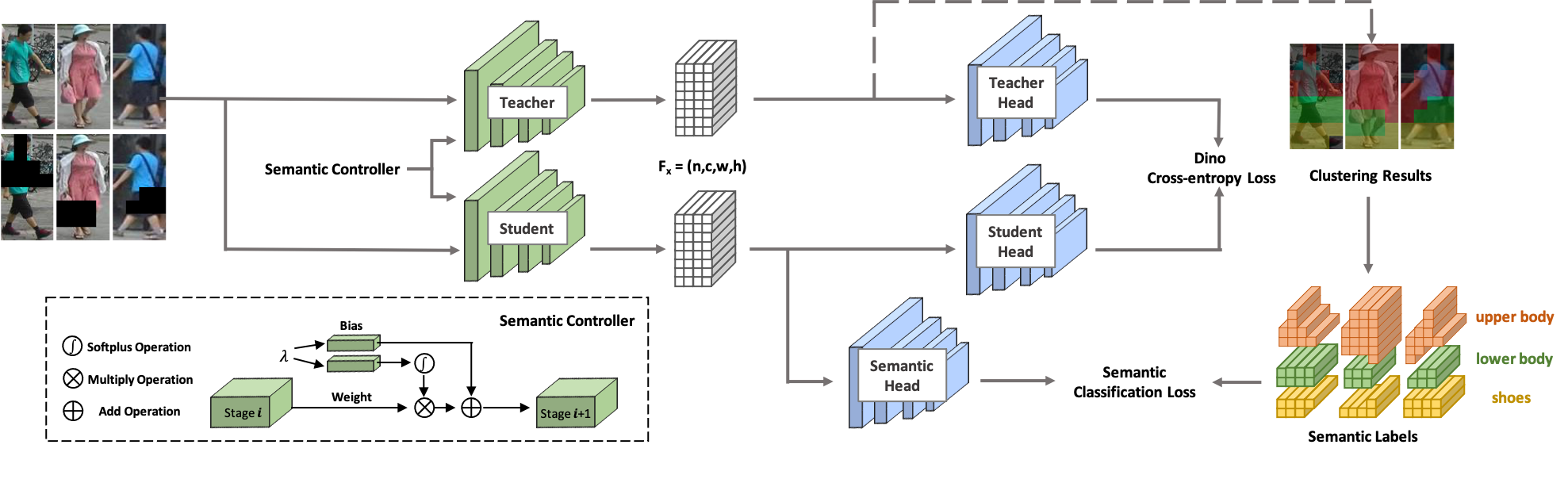}
\vspace{-1.1cm}
\caption{The pipeline of the proposed SOLIDER.}
\label{fig:framework}
\vspace{-0.4cm}
\end{figure*}

\section{Related Work}

\subsection{Self-supervised learning approaches}

Self-supervised learning approaches have attracted significant interests in computer vision, especially when used for learning image representation as the pretext task.

In current self-supervised learning, contrastive methods~\cite{moco,mocov2,swav,simclr,ibot,dino} have achieved great success, and provided state-of-the-art performance for image representation. The goal of contrastive learning is to minimize the distances between two augmented views of a same image~\cite{swav,simclr}, and distinguish each image from all the others~\cite{moco,mocov2,dino}. 
MoCo~\cite{moco,mocov2} improves the training of contrastive methods by storing representations from a momentum encoder instead of the trained
network. SimCLR~\cite{simclr} shows that the memory bank can be entirely replaced with the elements from the same batch if the batch is large enough. DINO~\cite{dino} combines most of these techniques, including momentum encoder~\cite{moco}, multi-crop training~\cite{swav}, and the use of small patches with ViTs~\cite{vit}, which builds a much better baseline.


Besides contrastive methods, masked image modeling methods~\cite{mae,simmim,beit} arouse extensive attention from researchers. 
BEiT~\cite{beit} proposes to predict discrete tokens. SimMIM~\cite{simmim} and MAE~\cite{mae}
find that a moderately large masked size and a light prediction head would be beneficial for the masked image modeling. 
Although masked language modeling manages to import semantic information into image representations, it can not explicitly figure out the semantic information from the image to supervise the training. In the proposed SOLIDER, we cluster tokens and use human prior to assign semantic labels to these tokens, which can train a stronger human semantic representation.

It is worth noting that DeepCluster~\cite{deepcluster} also use cluster to learn representations. But it is on image-level and can not produce semantic labels for tokens. Some unsupervised semantic segmentation methods~\cite{Cho2021PiCIE} extend DeepCluster on pixel-level. But it still faces problems and can not learn a satisfied representation, which is explained in Section.~\ref{sec:humanprior}

\subsection{Human-centric visual tasks}

By looking through human-centric visual tasks in computer vision, we find that there are many tasks directly or indirectly related to human\footnote{In this paper, we focus on feature representations for images, therefore, only downstream tasks on 2D image-level are our main concerns, and tasks about 3D data or sequential data are out of our scope.}, \eg, person re-identification~\cite{reidsurvey1,gu20201st,isobe2021towards,dou2022reliability,gu2022multi}, attribute recognition~\cite{attrsurvey1,attrsurvey2,pa100k}, person search~\cite{cuhk-sysu,prw}, pedestrian detection~\cite{detsurvey1,detsurvey2,pedestron}, multi-object tracking~\cite{chen2016tcsvt,chen2014novel,liu2021city,liu2022adaptive,tang2021simple}, human parsing~\cite{parsesurvey1,lip} and pose estimation~\cite{posesurvey1,posesurvey2}.

Among them, we sort out six representative tasks, 
\ie, person re-identification, attribute recognition, person search, pedestrian detection, human parsing and pose estimation. 
Person re-identification~\cite{jiang2021exploring,zhang2022graph,chen2022tagperson,wang2022refining,pan2022dynamic} aims at retrieving a person of interest across multiple non-overlapping cameras. 
Attribute recognition~\cite{attrsurvey1,peta,rap,pa100k} tries to mine the attributes of target people when given person images, of which the attributes are understandable semantic descriptions. 
Person search~\cite{cuhk-sysu,prw} aims to find a probe person from the whole scene which shows great significance in video surveillance community to track lost people.
Pedestrian detection~\cite{detsurvey1,pedestron,cityperson,caltech} focuses on detecting people in general images, which is a key ability for a variety of important applications. 
Human parsing~\cite{parsesurvey1,lip,schp}, a sub-task of semantic segmentation, aims to understand human-body parts on pixel level.
Human pose estimation~\cite{posesurvey1,mpii,hrformer} dedicates to locate the human body skeletons from images. 

Recently, Wang et al.~\cite{zeng2022not} also focus on human-centric visual tasks and design a Token Clustering Transformer (TCFormer), but TCFormer is a supervised method which can not take advantages of massive public unlabeled data.

\section{The SOLIDER}

The whole pipeline of the proposed SOLIDER is shown in Fig.~\ref{fig:framework}. In this section, we first explain how to generate pseudo semantic labels from human prior knowledge and  use it to supervise a token-level semantic classification pretext task. Then, we introduce how the learned representation can be controlled in SOLIDER.

\subsection{Semantic Supervision from Human Prior}
\label{sec:humanprior}

DINO~\cite{dino} is a state-of-the-art self-supervised learning method, which is widely used for image representation. As a constrastive learning based method, the visual appearance information is well learned in DINO's representation. We use DINO as our baseline and plan to involve more semantic information into its representation.

\begin{figure}[!t]
\centering
\includegraphics[width=1.0\linewidth]{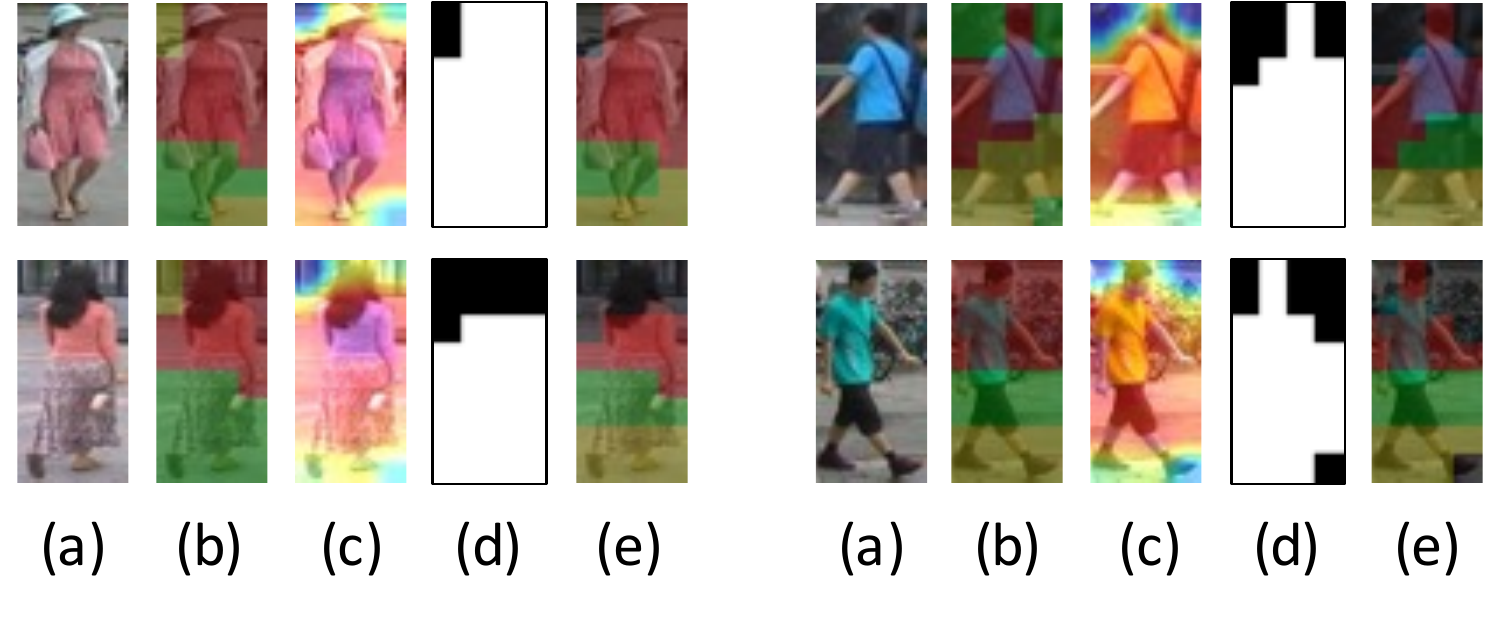}
\vspace{-0.8cm}
\caption{(a) The original images. (b) Initial semantic clustering results. (c) Attention maps. (d) Foreground clustering results. (e) Final semantic clustering results.}
\label{fig:cluster}
\vspace{-0.3cm}
\end{figure}

We cluster the token vectors from the learned DINO representation and show in Fig.~\ref{fig:cluster}(b). It can be seen that the representation is split into several parts based on its visual appearance. In other words, it can find things in images, although be not able to tell their meanings.

Some methods~\cite{deepcluster,Cho2021PiCIE} go further and try to assign semantic labels for these clustered things. Instead of cluster on single image, they do the clustering across images and aim to build the semantic relationship among images. However, as the features used for clustering are trained from visual appearance clue, the clustered results are dominated by appearance.
As shown in Fig.~\ref{fig:dino_space}, in original DINO space, the features of ``blue short'' (\textcolor{blue}{\textbf{+}}) is closer to ``blue shirt'' (\textcolor{red}{$\blacktriangle$}) due to similar appearance, but farther to ``black short'' (\textcolor{red}{\textbf{+}}), even though they share the same semantic meaning of ``short pants''. After clustering, the blue things (``blue short'' and ``blue shirt'') will be clustered together, and the things with the same semantic meaning (``blue short'' and ``black short'') are split apart. In other words, these methods can not produce the semantic labels we wanted.

After scanning through the unlabeled human images, we find that most of images have a fixed pattern~\cite{pcb,isp}: the person body erectly occupies the entire image, and the head is always on the top of the image while the feet is at the bottom of the image. With this observation, we give semantic labels to clustered parts of all images based on the order of their y-axis coordinates, \ie, the top part of all images are labeled as upper body, and the bottom part is marked as shoes. These pseudo semantic labels provide semantic information for every token vector. 
A token-level semantic classification pretext task is imported and supervised by these semantic labels. To better use these labels, we involve extra modifications, listed as following.

\textbf{Involving Background\&Foreground Clustering.} In these labels, we find that there are some background fragments which cause noise disturbance on the clustering results and mislead the alignment. To handle this problem, we introduce  another clustering before the semantic clustering. Specifically, we observe that the background tokens always have smaller responses compared to foreground tokens during training~\cite{isp,mauthner2015encoding}, as shown in Fig.~\ref{fig:cluster}(c). Thus, we cluster the token vectors into two categories based on their vector magnitudes, \ie, foreground and background. The results are shown in Fig.~\ref{fig:cluster}(d). Then the semantic clustering is only performed on the foreground tokens, and the new results are listed in Fig.~\ref{fig:cluster}(e). It can be clearly seen that the foreground tokens are well clustered into three semantic parts, \ie, upper body, lower body, shoes\footnote{In this paper, we treat heads/chests/arms as upper body, waist/thighs as lower body and calves/feet as shoes.}.

\textbf{Involving Masked Image Modeling.} Inspired by the masked image modeling methods~\cite{mae,simmim}, we would like to further introduce more semantic information into the representation. As we have known, people can easily locate every semantic part in a human image even when some parts are occluded. Therefore, we assume that if a semantic part is missing or occluded in the human image, the model would still be able to predict its semantic meaning based on other surrounding parts. For this purpose, we upgrade our semantic supervision into masked semantic supervision. Specifically, we randomly mask out a semantic part from the image $x$ and re-feed this masked image $\tilde{x}$ through the framework. Then the output tokens are supervised by the original semantic labels, because it is expected that the model would be able to provide the true semantic labels for the masked tokens with the help of other tokens.

The whole semantic self-supervision is presented in Fig.~\ref{fig:framework}.
Specifically, during training, for every iteration, we obtain the output feature maps $F$ from backbone with the size of ($n,c,h,w$). For each image $x$, we consider its feature maps\footnote{The feature maps $F_t$ from the teacher network is chosen for semantic clustering. The feature maps from the student network is marked as $F_s$.} $F_t$ as $w*h$ token vectors with the size of $c$. Then we use K-means~\cite{kmeans} to cluster them into two categories according to the magnitude (\ie, l2 normalization) of $c$, and consider the category with larger magnitude as foreground mask $M$. After that, another K-means is applied on the tokens in foreground mask $M$ to conduct the semantic clustering, to obtain $N$ predefined semantic categories
and assign semantic label $y$ for each token. Meanwhile, a semantic head\footnote{The semantic head contains several blocks, and each block includes a fully connected layer, a batch norm layer and a ReLU. } is involved to classify the vectors $F_s$ from the student branch based on these semantic labels $y$. The corresponding semantic classification loss is as below:

\begin{equation}
L_{sm}=\frac{1}{w\times h}\sum_{u\in w\atop v\in h}\sum_{i=1}^{N+1}-y^{(u,v)}\log\frac{f_s(u,v)^{(i)}}{\sum_{k=1}^{N+1}f_s(u,v)^{(k)}}
\label{eq:scloss}
\end{equation}
where $f_s\!=\!h_{sm}(flatten(F_s))$. $flatten()$ is used to reshape $F_s$ from ($n,c,h,w$) to ($n*h*w,c$), and $h_{sm}$ is the semantic head. $N$ means the number of clustered semantic parts. $f_s(u,v)^{(i)}$ indicates the predicted probability of token $(u,v)$ on part $i$. After image $x$, we randomly mask out a part of $x$ to obtain image $\tilde{x}$, and send $\tilde{x}$ to Eq.~\ref{eq:scloss} too. 

The total loss for the whole SOLIDER framework is:
\begin{equation}
L = \alpha L_{dino} + (1-\alpha)L_{sm}
\label{eq:allloss}
\end{equation}
where $\alpha$ is a balance weight and set to 0.5 by experience.

\subsection{Semantic Controller}
\label{ssec:control}

When the learned representation is used as the pre-trained model for downstream tasks, we expect the appearance and semantic information learned in the representation can help downstream tasks. However, each downstream task has its own focus. For example, in person re-identification~\cite{reidsurvey1,reidsurvey2}, although the semantic information can help to align parts across person images~\cite{suh2018part,zhang2017alignedreid}, the appearance information is the key clue to distinguish different people, which is the most important information~\cite{yang2016metric,zheng2019joint}. So a pre-trained model with more appearance information and less semantic information would provide a better startup for person re-identification task. On the contrary, in pedestrian detection~\cite{detsurvey1,detsurvey2} and human parsing~\cite{parsesurvey1,lip,schp}, the semantic information plays a key role for these tasks, while the appearance difference is useless and should be partly ignored. Its pre-trained model is expected to contain enough semantic information. 

To fit the needs of different tasks, it requires the pertained model can be adjusted according to the downstream tasks. However, as the pertained model is trained, it is hard to change its parameters for different downstream tasks. Task token~\cite{tasktoken} is a potential way to solve this problem, which pre-sets an extra one-hot token for each task. 
But it has several problems. First, the number of task tokens should be pre-defined before learning representation. In real applications, we can not determine what tasks the learned representation would be utilized on in advance. Second, although we can pre-defined task tokens from semantic perspective (two task tokens, \ie, with and without semantic information), the task token is a discrete variable, which leaves the downstream tasks with limited choice of representations. Third, the task token is essentially a re-weighted sum of original learned tokens. It is hard to cooperate with Swin~\cite{swin} backbone, which is a more friendly transformer backbone for visual tasks, because of the shifted windows in Swin~\cite{swin}.

In this paper, SOLIDER uses a conditional network with a semantic controller to handle this problem. 
During the pretext task training, due to the unavailability of downstream task, we can not adjust the representation for specific task, so we import the semantic controller to make the pre-trained model conditional. 
The inputs of the semantic controller are the image feature maps and a continuous value $\lambda \in [0,1]$. The  $\lambda$ represents the required ratio of semantic information in representation. The output is the new feature maps with the ratio we required. The details of the proposed semantic controller can be seen in Fig.~\ref{fig:framework}. In semantic controller the value $\lambda$ is encoded into a weight vector and a bias vector. After a Softplus activation function, the weight vector is multiplied onto the original feature maps, and the bias vector is added for final outputs. 

We apply the semantic controller after each block of Swin Backbone, and the feature maps $F(\lambda)$ with new ratio $\lambda$ is sent to the next block. We use the following equation to produce our semantic controllable model: 
\begin{equation}
L = \alpha L_{dino}(F(\lambda)) + \lambda(1-\alpha)L_{sm}(F(\lambda))
\label{eq:newallloss}
\end{equation}
In the pretext task training stage, $\lambda$ is randomly sampled for every iteration. We tried different distributions of $\lambda$, \textit{i.e.}, binomial distribution $B(p\!=\!0.5)$, continuous uniform distribution $U[0,1]$ and beta distribution $\beta(0.2,0.2)$. $\beta(0.2,0.2)$ is better than $U[0,1]$, implying that emphasizing the sampling on two borders is more important for training the controller, which is consistent with the conclusions of Mixup~\cite{zhang2017mixup}. Finally, $B(p\!\!=\!\!0.5)$ is found to perform the best, \textit{a.k.a.}, a binary distribution from \{0,1\}. 
When the pre-trained model is applied to downstream tasks, $\lambda$ is set manually to adapt the pre-trained model to each downstream task. The pre-trained model with $\lambda$ provides a better startup for downstream fine-tuning, and all parameters of the pre-trained model are fine-tuned. 

In sum, after a conditional pre-trained model is trained from SOLIDER, we can send different  $\lambda$ to the model to produce representations with different ratios of semantic information for downstream tasks. 
A training pseudo code is provided in Supplementary Material to further clarify the whole training process of SOLIDER.

\section{Experiments}
The pre-trained model from SOLIDER is verified on six downstream human-centric visual tasks, including person re-identification, attribute recognition, person search, pedestrian detection, human parsing and pose estimation. 

\begin{figure*}[!t]
\centering
\includegraphics[width=1.0\linewidth]{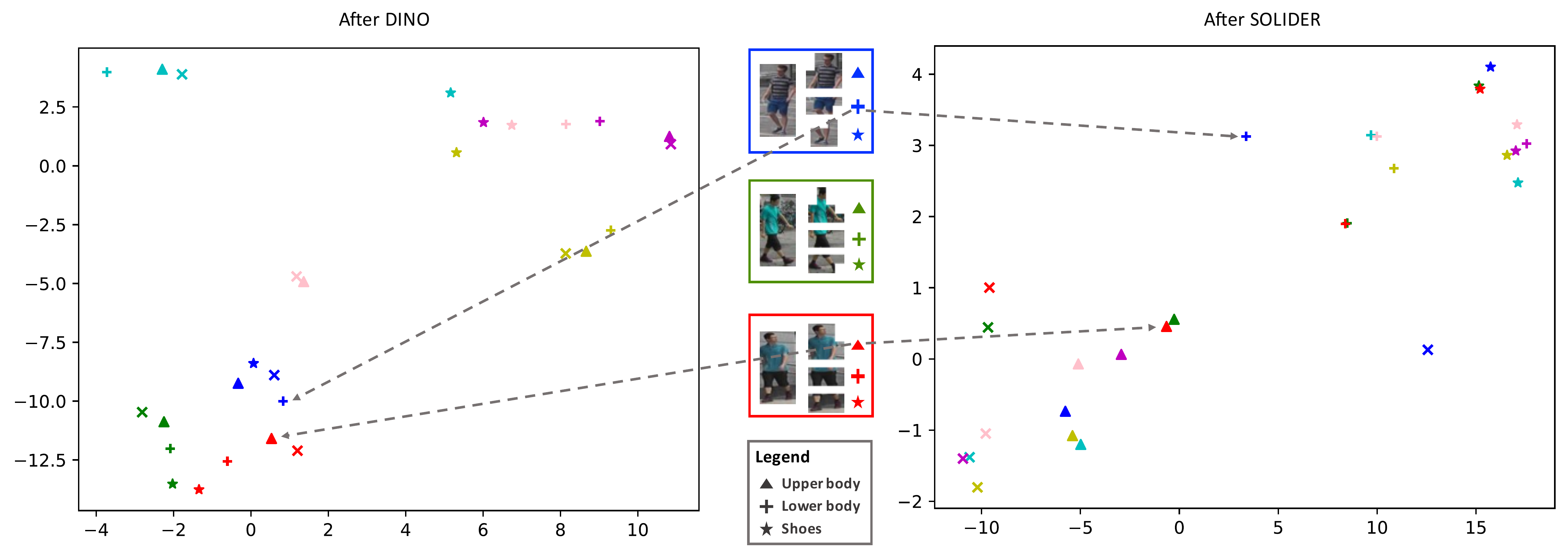}
\vspace{-0.7cm}
\caption{The representation space learned before and after involving SOLIDER. For each image, we split its features into four parts according to their semantic regions, \ie, upper body (as $\blacktriangle$), lower body (as \textbf{+}), shoes (as $\star$) and background (as \textbf{$\times$}, not shown for a clearer presentation). It can be seen that before SOLIDER is introduced, the features tend to gather by appearance. While after SOLIDER, the features with same semantic meanings are closer to each other.}
\label{fig:visual}
\vspace{-0.3cm}
\end{figure*}

\begin{figure}[!t]
\centering
\includegraphics[width=0.95\linewidth]{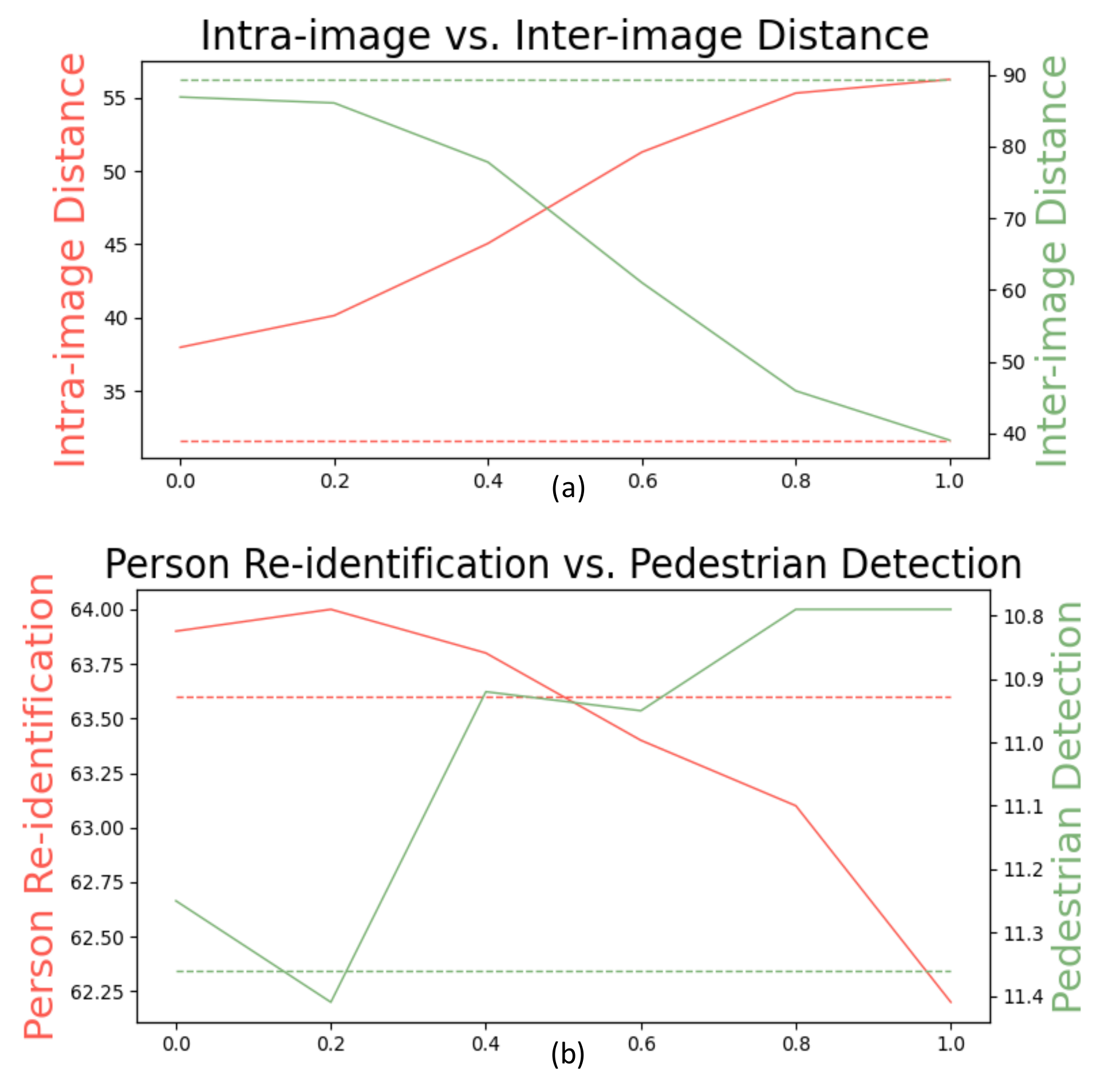}
\vspace{-0.4cm}
\caption{(a) The intra-image and intra-image distances under different $\lambda$. (b)The performance of person re-identification and pedestrian detection under different $\lambda$. The dash line indicates two distances and two task results from the original DINO model.}.
\label{fig:analysis}
\vspace{-0.6cm}
\end{figure}

\subsection{Training Settings}

\subsubsection{Datasets} 
For pretext tasks, LUPerson~\cite{lup,fu2022large} is used for training, the same as~\cite{luo2021self,pass}. It contains 4.18M human images without any label. From each downstream task, we conduct experiments on their commonly-used datasets. 
Specifically, in person re-identification, the experiments are conducted on Market1501~\cite{market2015iccv} and MSMT17~\cite{msmt17}. 
In attribute recognition, PETA$_{zs}$~\cite{attreval}, RAP$_{zs}$~\cite{attreval} and PA100k~\cite{pa100k} are considered. 
In person search, we adopt CUHK-SYSU~\cite{cuhk-sysu} and PRW~\cite{prw} in our experiments. 
CityPerson~\cite{cityperson} is utilized for pedestrian detection. 
In human parsing and pose estimation, LIP~\cite{lip} and COCO~\cite{coco} pose estimation are used respectively. 

\subsubsection{Evaluation Metric} 
In person re-identification and person search, mAP/Rank1~\cite{market2015iccv} are adopted as evaluation metrics. 
For attribute recognition, the evaluation metrics are mean accuracy (mA)~\cite{attreval}. 
To evaluate the performance on pedestrian detection, we employ the log-average Miss Rate over false positive on Reasonable and Heavy Occluded (MR$^{-2}$ on R/HO)~\cite{pedestron}.
Human parsing uses mIoU~\cite{schp} for evaluation, and pose estimation utilizes Average Precision/Recall (AP/AR)~\cite{hrformer} as the evaluation metric.

\subsubsection{Optimization} 
We use Swin-Transformer~\cite{swin} as the backbone throughout all the experiments. 
In the pretext task training, 
SGD is used as the optimizer, and the learning rate is 0.0005. The model is trained for 100 epochs, and the learning rate is declined by Cosine Annealing scheduler. The batch size is 48/32/24 when training on  8 Tesla V100 32G GPUs for Swin-Tiny/Small/Base.

It is worth noting that, as the semantic clustering process is time-consuming, we do not directly train SOLIDER with 100 epochs from scratch. Instead, we first train DINO with 100 epochs, and then finetune the SOLIDER on the trained DINO model with another 10 epochs using a smaller learning rate 0.00005.

For downstream tasks, we reproduce a state-of-the-art method in each task as our baseline. The methods we used as baseline are listed in Table.~\ref{tab:abl}. What we do is just replacing their backbones to the Swin-Transformer~\cite{swin} backbone which is pre-trained by the proposed SOLIDER.

\subsection{Qualitative and Quantitative Analysis}

\textbf{Analysis on semantic clustering.} To observe the semantic representation ability of our pre-trained model in qualitative analysis, we randomly select some images from the training data, and visualize their representation features learned before and after SOLIDER. The visualization is provided in Fig.~\ref{fig:visual}. It can be found that, before introducing the semantic supervision, the representation features are distributed mainly based on the identities of the images.
And the images sharing similar appearance stay closer, even with different semantic meanings. 
For example, the features of ``blue short'' (\textcolor{blue}{\textbf{+}}) is closer to ``blue shirt'' (\textcolor{red}{$\blacktriangle$}) due to similar appearance ``color blue'', but further to ``black short'' (\textcolor{red}{\textbf{+}}), even they share the same semantic meanings ``short pants''.

However, after involving the semantic supervision by SOLIDER, the feature distribution pays more attention on the semantic meaning. We can find that the images hold similar semantic meanings are closer to each other, even though they share different appearance. 
It implies that the human prior discovered by our SOLIDER can provide a better semantic supervision for representation, which helps the pre-trained model involves more semantic meanings.

\begin{table*}[!t]
\renewcommand\arraystretch{1.8}
\caption{The performance of different pre-trained models on downstream tasks. ``Sup'' implies the supervised training. ``+Clustering'' indicate the training with semantic supervision. ``+Control'' means the semantic controller is involved. For each downstream task, we list its name, evaluation metric and the state of the art that we used as our baseline. $\uparrow$/$\downarrow$ means the larger/smaller value the better performance.}
\vspace{-0.6cm}
\begin{center}
\scalebox{0.85}{
\begin{tabular}{|c|c|c|c|c|c|}
\hline
\multicolumn{2}{|c|}{Pretrain Methods} & Sup & DINO~\cite{dino} & \makecell[c]{+ Clustering} & \makecell[c]{+ Clustering\&Controller} \\
\hline
\multicolumn{2}{|c|}{Pretrain Data} & ImageNet & LUP1M & LUP1M & LUP1M \\
\hline
\specialrule{0em}{2pt}{0pt}
\hline
\multirow{2}{*}{\makecell[c]{\textbf{Person Re-identification}\\\footnotesize{mAP/Rank1 $\uparrow$}\\TransReID~\cite{transreid}}} & Market1501 & 78.1/90.2 & 89.6/95.9 & 89.5/95.5 & \textbf{89.9/96.1}\\
\cline{2-6}
& MSMT17 & 49.7/73.6 & 63.3/83.2 & 61.6/82.2 & \textbf{63.9/83.8} \\
\hline
\specialrule{0em}{2pt}{0pt}
\hline
\multirow{3}{*}{\makecell[c]{\textbf{Attribute Recognition}\\\footnotesize{mA $\uparrow$}\\RethinkPAR~\cite{attreval}}} & PETA$_{zs}$ & 72.86 & 73.64 & 73.90 & \textbf{74.20}\\
\cline{2-6}
 & RAP$_{zs}$ & 72.10 & 73.04 & 73.16 & \textbf{73.21} \\
\cline{2-6}
 & PA100k & 80.67 & 82.98 & 82.98 & \textbf{84.15} \\
\hline
\specialrule{0em}{2pt}{0pt}
\hline
\multirow{2}{*}{\makecell[c]{\textbf{Person Search}\\\footnotesize{mAP/Rank1 $\uparrow$}\\SeqNet~\cite{seqnet}}} & CUHK-SYSU & 93.0/94.1 & 93.6/94.3 & 93.6/94.1 & \textbf{94.0/94.7} \\
\cline{2-6}
 & PRW & 50.0/84.4 & 52.9/84.7 & 53.0/84.0 & \textbf{54.1/85.0} \\
\hline
\specialrule{0em}{2pt}{0pt}
\hline
\makecell[c]{\textbf{Pedestrian Detection}\\\footnotesize{MR$^{-2}$(R/HO) $\downarrow$}\\CSP~\cite{csp}} & CityPerson & 11.6/43.8 & 11.4/43.1 & 11.1/41.7 & \textbf{10.8/40.7}\\
\hline
\specialrule{0em}{2pt}{0pt}
\hline
\makecell[c]{\textbf{Human Parsing}\\\footnotesize{mIOU} $\uparrow$\\SCHP~\cite{schp}} & LIP & 51.10& 54.45& 55.25& \textbf{55.45} \\
\hline
\specialrule{0em}{2pt}{0pt}
\hline
\makecell[c]{\textbf{Pose Estimation}\\\footnotesize{AP/AR} $\uparrow$\\HRFormer~\cite{hrformer}} & COCO & 72.4/78.2 & 73.1/78.5 & 73.4/78.7 & \textbf{74.4/79.7}\\
\hline
\end{tabular}}
\end{center}
\label{tab:abl}
\vspace{-0.3cm}
\end{table*}

\textbf{Analysis on semantic controller.} To further verify the semantic controller, we provide another two experiments. We first define two distance, \ie, intra-image distance and inter-image distance. Intra-image distance is the average distance between any two parts from the same image. Inter-image distance indicates the average distance between parts of the same semantic meaning but coming from different images. Small intra-image distance and large inter-image distance indicate the appearance information is dominant in the representation, otherwise the semantic information dominates. After the SOLIDER is trained, we set the semantic weight $\lambda$ from 0 to 1, to observe two distances of these different representations in the whole LUP dataset. The results are shown in Fig.~\ref{fig:analysis}(a). It can be seen that with the increase of $\lambda$, the intra-image distance gets larger, while the inter-image distance becomes smaller, which suggests that more semantic information is involved into the representation and become dominant.

Meanwhile, as we've known, person re-identification requires more appearance information, while pedestrian detection is prone to the semantic information. We list the performance of two tasks using the pre-trained representations with different $\lambda$ As shown in Fig.~\ref{fig:analysis}(b), with the increase of $\lambda$, the person re-identification performance becomes worse and the pedestrian detection performance turns to better. In other words, the representation with a large $\lambda$ can provide a better startup for pedestrian detection, while a small $\lambda$ fits more to person re-identification.
This phenomenon further implies the effectiveness of our semantic controller.

\subsection{Ablation Study}
\label{ssec:ablation}

The ablation study is provide to verify the effectiveness of each module in SOLIDER, and all the experiments are conducted on Swin-Tiny~\cite{swin} backbone. We use different methods to train the pretext task, and verify pre-trained models on all downstream human-centric visual tasks.

ImageNet and LUP1M are datasets used for the pretext task training. 
ImageNet indicates the model is trained on ImageNet. This training is supervised with labels in ImageNet. LUP1M is a subset randomly sampled from LUP, which contains 1 million person images and has a similar image number with ImageNet for fair comparison. Due to no labels in LUP1M, the training process is self-supervised. We take DINO~\cite{dino} as our baseline. ``+Clustering'' indicates that we involve the human prior into DINO by semantic clustering. ``+Clustering\&Controller'' means that the semantic controller is imported to control the ratio of semantic information in representation. In downstream tasks, we set $\lambda$ to a small value (0.0-0.2) for person-re-identification, and a large value (0.8-1.0) for pedestrian detection, human attributes, human parsing and pose estimation. For person search, where appearance information is as important as semantic information, we set  $\lambda$ to a moderate value. Details of the $\lambda$ selection for downstream tasks are provided in Supplementary Material.

From the comparison in Table.~\ref{tab:abl}, 
it can be found that after involving the semantic information (``+Clustering''), the pre-trained model achieves a better performance on most of downstream tasks. It implies the effectiveness of the representation with more semantic information. It is worth noticing that the performance on person re-identification and person search are somewhat declined after involving semantic supervision. The reason is that after the semantic information is imported, the model is more inclined to represent semantic information, and its power on distinguishing different identities is weakened to some extent, which needs a more careful balance. The semantic controller is exactly designed for this problem. After involving the semantic control (``+Controller''), the performance further improved, especially for person re-identification and person search. 
Comparing our SOLIDER results to DINO's on person re-identification, we can observe that our improvement is limited. Because the key clue in person re-identification is appearance information, which is already well learned in DINO. The semantic information imported from SOLIDER plays an auxiliary role leading to only a slight increment.

\begin{table*}[!t]
\renewcommand\arraystretch{1.5}
\setlength\tabcolsep{1.9pt}
\caption{The comparison of the proposed SOLIDER with other state of the arts.}
\vspace{-0.6cm}
\begin{center}
\scalebox{0.8}{
\begin{tabular}{|c|c|c|c|c|c|c||ccc|}
\hline
\multirow{3}{*}{\shortstack{\textbf{\makecell[c]{Person\\Re-identification}}\\\footnotesize{mAP/Rank1 $\uparrow$}}} &  & SCSN~\cite{chen2020salience} & ABDNet~\cite{chen2019abd} & TransReID~\cite{transreid} & UP-ReID~\cite{yang2022unleashing} & PASS~\cite{pass} & \makecell[c]{\\Swin-T} & \makecell[c]{SOLIDER\\Swin-S} & \makecell[c]{\\Swin-B}\\
\cline{2-10}
 & Market1501 & 88.5/95.7 & 88.3/95.6 & 89.5/95.2 & 91.1/97.1 & 93.3/\textbf{96.9} & 91.6/96.1 & 93.3/96.6 & \textbf{93.9/96.9} \\
\cline{2-10}
 & MSMT17 & 58.5/83.8 & 60.8/82.3 & 69.4/86.2 & 63.3/84.3 & 74.3/89.7 & 67.4/85.9 & 76.9/90.8 & \textbf{77.1/90.7} \\
\hline
\specialrule{0em}{2pt}{0pt}
\hline
\multirow{4}{*}{\shortstack{\textbf{\makecell[c]{Attribute\\Recognition}}\\\footnotesize{mA $\uparrow$}}} &  & MsVAA~\cite{msavv} & VAC~\cite{vac} & ALM~\cite{alm} & JLAC~\cite{jlac} & RethinkPAR~\cite{attreval} & \makecell[c]{\\Swin-T} & \makecell[c]{SOLIDER\\Swin-S} & \makecell[c]{\\Swin-B} \\
\cline{2-10}
 & PETA$_{zs}$ & 71.53 & 71.91 & 73.01 & 73.60 & 71.62& 74.37 & 76.21 & \textbf{76.43}\\
\cline{2-10}
 & RAP$_{zs}$ & 72.04 & 73.70 & 74.28 & 76.38 & 72.32 & 74.23 & 76.84 & \textbf{77.06}\\
\cline{2-10}
 & PA100k & 80.41 & 79.16 & 80.68 & 82.31 & 81.61 & 84.14 & 86.25 & \textbf{86.37}\\
\hline
\specialrule{0em}{2pt}{0pt}
\hline
\multirow{3}{*}{\shortstack{\textbf{\makecell[c]{Person\\Search}}\\\footnotesize{mAP/Rank1 $\uparrow$}}} & & NAE+~\cite{nae} & AlignPS+~\cite{alignps} & TCTS~\cite{tcts} & SeqNet~\cite{seqnet} & GLCNet~\cite{glcnet} & \makecell[c]{\\Swin-T} & \makecell[c]{SOLIDER\\Swin-S} & \makecell[c]{\\Swin-B} \\
\cline{2-10}
 & CUHK-SYSU & 92.1/92.9 & 94.0/94.5 & 93.9/95.1 & 94.8/95.7 & \textbf{95.8/96.2} & 94.9/95.7 & 95.5/95.8 & 94.9/95.5\\
\cline{2-10}
 & PRW & 44.0/81.1 & 46.1/82.1 & 46.8/87.5 & 47.6/87.6 & 47.8/\textbf{87.8} & 56.8/86.8 & \textbf{59.8}/86.7 & 59.7/86.8 \\
\hline
\specialrule{0em}{2pt}{0pt}
\hline
\multirow{2}{*}{\shortstack{\textbf{\makecell[c]{Pedestrian\\Detection}}\\\footnotesize{MR$^{-2}$(R/HO) $\downarrow$}}} &  & RepLoss~\cite{reploss} & CSP~\cite{csp} & NMS-Loss~\cite{nmsloss} & ACSP~\cite{acsp} & PedesFormer~\cite{pedesformer} & \makecell[c]{\\Swin-T} & \makecell[c]{SOLIDER\\Swin-S} & \makecell[c]{\\Swin-B} \\
\cline{2-10}
 & CityPerson & 13.2/56.9 & 11.0/49.3 & 10.8/- & 9.3/46.3 & \textbf{9.2/36.9} & 10.3/40.8 & 10.0/39.2 & 9.7/39.4 \\
\hline
\specialrule{0em}{2pt}{0pt}
\hline
\multirow{2}{*}{\shortstack{\textbf{\makecell[c]{Human\\Parsing}}\\\footnotesize{mIOU} $\uparrow$}} &  & JPPNet~\cite{liang2018look} & BraidNet~\cite{liu2019braidnet} & CE2P~\cite{ruan2019devil} & PCNet~\cite{zhang2020part} & SCHP~\cite{schp} & \makecell[c]{\\Swin-T} & \makecell[c]{SOLIDER\\Swin-S} & \makecell[c]{\\Swin-B} \\
\cline{2-10}
 & LIP &51.37 &54.40 & 53.10 & 57.03 & 59.36 & 57.52 & 60.21 & \textbf{60.50	} \\
\hline
\specialrule{0em}{2pt}{0pt}
\hline
\multirow{2}{*}{\shortstack{\textbf{\makecell[c]{Pose\\Estimation}}\\\footnotesize{AP/AR} $\uparrow$}} &  &  CPN~\cite{chen2018cascaded} & SimpleBase~\cite{xiao2018simple} & TokenPose~\cite{li2021tokenpose} & HRNet~\cite{sun2019deep}& HRFormer~\cite{hrformer} & \makecell[c]{\\Swin-T} & \makecell[c]{SOLIDER\\Swin-S} & \makecell[c]{\\Swin-B} \\
\cline{2-10}
 & COCO & 68.6/- & 74.3/79.7 & 75.8/80.9 & 76.3/81.2& \textbf{77.2/82.0} & 74.4/79.6 & 76.3/81.3 & 76.6/81.5\\
\hline
\end{tabular}}
\end{center}
\label{tab:sota}
\vspace{-0.5cm}
\end{table*}

It is noteworthy that the larger improvement of DINO LUP1M compared to Sup ImageNet on person ReID is due to the street-view scenarios of LUPerson images which are similar to ReID datasets. Except for the ReID task, the DINO LUP1M brings an average of 1.4 improvement (72.6 vs. 74.0) on five tasks compared to Sup ImageNet, which implies the advantage of the appearance information learned from DINO. With SOLIDER, the average performance is raised to 74.9. The further 0.9 increase shows the success of involving semantic information by our SOLIDER. And this improvement is consistent on all tasks.

Besides, we conduct experiments on the influence of the clustered part number and the semantic head size in SOLIDER, which are present in Supplementary Material. 

\subsection{Comparison to State of the Arts}

We compare our results with state of the arts on six human-centric tasks. The state-of-the-art methods for comparison are listed in Table.~\ref{tab:sota}. 
In person re-identification, we use TransReID without side information as our baseline. Even without side information, we can achieve a better performance than other self-supervised works, \eg, TransReID~\cite{transreid} and PASS~\cite{pass}, which also trained on LUPerson dataset.
For the PRW results of person search, the mAP of our results  outperform other state of the arts with more than 10\%. The mAP criterion reflects the detection ability of models. It implies that our pre-trained model can lead to a better detection result, which may thanks to the imported semantic information.
In pedestrian detection, PedesFormer~\cite{pedesformer} achieves a better performance than ours on pedestrian detection task is because it also involves extra data from autonomous driving datasets which are specific for pedestrian detection task and not for other human-centric tasks. 
For pose estimation, we report the best performance of HRFormer~\cite{hrformer} which is trained on HRFormer backbone instead of Swin. 
After HRFormer~\cite{hrformer} backbone is switched to Swin, the result is 75.9/81.1 reported in MMPose~\cite{mmpose2020}, which is lower than ours 76.6/81.5.
From the comparison in Table.~\ref{tab:sota}, we can see that the pre-trained from SOLIDER can provide a better initialization for these human-centric tasks, and can be used as a new baseline for further works on these tasks. 

We also conduct SOLIDER on different backbones, also summarize in Table.~\ref{tab:sota}.
It can be seen that with the model size increasing in Swin-Transformer backbones, the performance is further improved. In some tasks, Swin-Small achieves a better performance than Swin-Base, it is because of a larger batch size in Swin-Small compared to that in Swin-Base under a limited GPU memory.

\section{Conclusion}
This paper proposes a semantic controllable self-supervised learning framework called SOLIDER. It can utilize prior knowledge from human images to train representations with more semantic information. Moreover, the pre-trained model from SOLIDER can be adjusted by an input  value through the semantic controller, which can produce representations with different ratios of semantic information and to satisfy the requirements on downstream tasks. The human representations from SOLIDER is verified on six human-centric visual tasks, which can promote the development of these human-centric tasks in computer vision community.

{\small
\bibliographystyle{ieee_fullname}
\bibliography{egbib}
}

\end{document}